\documentclass[letterpaper, 10pt, conference]{ieeeconf}

\IEEEoverridecommandlockouts
\usepackage{graphicx}
\usepackage[caption=false,font=footnotesize]{subfig}
\usepackage{amsfonts}
\usepackage{booktabs}
\usepackage{amsmath}
\usepackage{cite}
\usepackage{url}
\usepackage[font=small]{caption}

\title{\LARGE \bf
Implicit Behavior Coordination from Sub-Task Demonstrations by Exploiting Overlap-Induced Multimodality
\author{
  Ahmed Shokry\textsuperscript{1,2} \enspace
  Usama Ahmed Siddiquie\textsuperscript{1} \enspace
  Sicong Pan\textsuperscript{1,2} \enspace
  Maren Bennewitz\textsuperscript{1,2}\\[2mm]
  \thanks{$^1$Humanoid Robots Lab and Center for Robotics, University of Bonn, Germany. $^2$Lamarr Institute for Machine Learning and Artificial Intelligence, Bonn, Germany.  }
}
}

\begin{document}
\maketitle


\begin{abstract}

Long-horizon robotic rearrangement is commonly formulated as a skill-sequencing problem, where distinct behaviors are explicitly represented and coordinated by a planner or high-level policy.
We investigate whether such explicit behavior identities and sequencing interfaces are necessary at all.
We introduce implicit behavior coordination from sub-task demonstrations, where separately collected behaviors are coordinated without behavior identity labels, complete-task demonstrations, or task-ordering supervision.
Our key observation is that overlap between sub-task demonstrations induces multimodal action distributions that need not be resolved through explicit behavior partitioning.
Instead, this overlap-induced multimodality can be exploited as a coordination resource.
We instantiate this idea with a shared Flow Matching policy that preserves multiple action modes and critic-guided in-sample planning that propagates task value across demonstrations and selects task-relevant modes.
Experiments in Habitat and on a real robot show that implicit behavior coordination remains effective under reduced cross-behavior overlap, larger behavior mixtures, longer horizons, and execution failures, supporting the idea that long-horizon coordination can emerge directly from sub-task demonstrations without explicitly recovering or sequencing behavior identities.

\end{abstract}

\section{Introduction}
\label{sec:introduction}
Long-horizon robotic rearrangement is often formulated as a skill-sequencing problem \cite{M3,say_can}.
A robot is assumed to possess a set of distinct behaviors, such as navigation, object interaction, picking, and placing, while a planner or high-level policy decides which behavior to execute at each stage of the task.
This formulation has led to effective systems based on modular skill libraries, task planners, hierarchical policies, and language-conditioned coordinators \cite{task_planning_ex,asc,qiu2025wildlma}.
However, these systems retain an explicit coordination interface: behavior identities must be defined or inferred, transitions between behaviors must be represented, and a separate mechanism is required to select or sequence them.
This raises a more fundamental question: must behavior identity be explicitly represented at all for long-horizon coordination?

\begin{figure}[t]
\centerline{\includegraphics[width=0.9\linewidth]{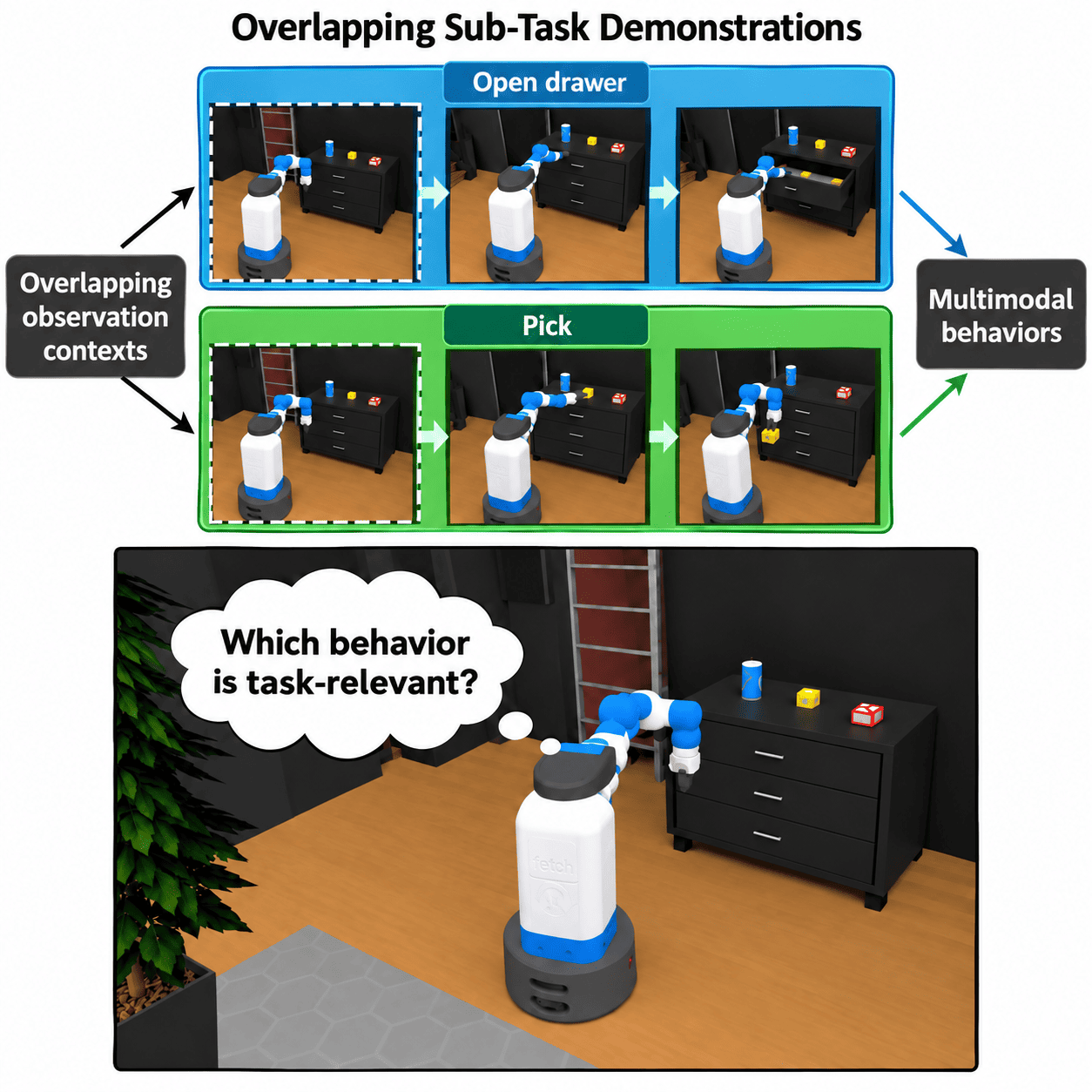}}
\caption{
Similar observation contexts can occur across demonstrations of different behaviors and prescribe different continuations.
This overlap-induced multimodality gives rise to a coordination problem: the agent must determine which plausible behavior continuation is relevant to the long-horizon task.
We address this by using a Flow Matching policy to preserve the multimodal behavior continuations and value-guided selection to choose the continuation with the highest expected return for the current task objective.
}
\label{fig:cover_fig}
\vspace{-0.5cm}
\end{figure}

In this work, we investigate whether long-horizon coordination requires such an explicit behavior abstraction at all.
We consider separately collected sub-task demonstrations, where each trajectory contains a useful behavior but the learner is not given its behavior identity, a complete-task trajectory, or annotations specifying how different behaviors should be ordered.
This setting creates a distinctive challenge.
Different sub-task demonstrations can visit similar observation contexts while prescribing different actions, inducing a multimodal action distribution, as illustrated in Fig.~\ref{fig:cover_fig}.
Existing approaches typically handle such multimodality in one of two ways: they either resolve it through explicit or latent behavior abstractions~\cite{M3,say_can,buds,autocgp}, or preserve it within an expressive generative policy for action modeling~\cite{chi2025diffusion,chisari2024learning}.
In these approaches, however, overlap itself is not used as an interface for cross-behavior coordination.

We take a different view: overlap-induced multimodality can itself be exploited as a resource for coordination.
At overlapping contexts, a shared generative policy can preserve multiple plausible action modes corresponding to different skill-like behaviors.
Task value can then propagate across separately collected demonstrations through these shared contexts, while the learned critic selects among the available modes according to the long-horizon objective.
In this way, behavior identities and transitions remain implicit rather than being recovered as an explicit skill variable or sequencing policy.
We instantiate this idea with a shared Flow Matching policy that preserves overlap-induced action modes, and uncertainty-aware in-sample planning that propagates task value across demonstrations while reducing the influence of unreliable value estimates.
Our contributions are threefold:
\begin{itemize}
    \item \emph{Implicit behavior coordination from sub-task demonstrations}: a learning formulation for long-horizon rearrangement that coordinates separately collected behaviors without behavior identity labels, complete-task demonstrations, or an explicit sequencing policy.
    \item \emph{Overlap-induced multimodality as a coordination resource}: an uncertainty-aware value propagation mechanism that reliably exploits multimodal behavior alternatives across demonstrations.
    \item \emph{Robustness and flexibility of implicit behavior coordination}: effective coordination despite suboptimal demonstrations and reduced cross-behavior overlap, and across larger mixed behavior repertoires, longer chained objectives, and real-world failures and recovery.
\end{itemize}


\section{Related Work}

\subsubsection{Explicit skill coordination for long-horizon mobile manipulation}
A common approach to long-horizon mobile manipulation is to decompose the task into individual navigation and manipulation skills and coordinate them using either a task planner~\cite{M3} or a learned high-level policy, such as Skill Transformer~\cite{skill_transformer} and ASC~\cite{asc}.
More recently, VLMs and LLMs have also been used for high-level coordination:
SayCan~\cite{say_can} uses an LLM to select among a library of pretrained skills based on language and affordance scores,
and FALCON~\cite{falcon} employs a foundation-model-based coordinator over decoupled locomotion and manipulation policies.
While these methods differ in how behaviors are represented and selected, they retain an explicit decomposition into skill or policy modules together with a dedicated coordination mechanism.
In contrast, we investigate whether behavior identities need to be explicitly represented at all for long-horizon coordination.

\subsubsection{Learning latent skills and abstractions from demonstrations}
A complementary line of work reduces the need for manually specified skill labels by discovering reusable abstractions directly from demonstrations.
BUDS~\cite{buds} discovers reusable skills from unsegmented robot demonstrations and composes them using a meta-controller, while AutoCGP~\cite{autocgp} automatically learns manipulation concepts from unlabeled demonstrations and explicitly selects among them for long-horizon execution.
These approaches show that meaningful behavior abstractions can be discovered without manual skill annotation.
However, the discovered skill or concept is still explicitly represented and used as a coordination variable.
In contrast, our method keeps behavior identity implicit and coordinates behavior modes directly through task value.

\subsubsection{Multimodal generative policies for robot actions}
Generative policies provide expressive models for multimodal action distributions in robot learning.
Diffusion Policy~\cite{chi2025diffusion} models multimodal visuomotor actions through conditional diffusion, while recent work extends generative policies to mobile manipulation~\cite{AC_DIT}.
Flow Matching has also been applied to robot manipulation~\cite{chisari2024learning}.
These works demonstrate the expressiveness of generative action models, including their ability to represent multimodal action distributions.
In contrast, we exploit overlap-induced multimodality across sub-task demonstrations for value-guided coordination with a shared generative policy.

\subsubsection{Value-guided policy learning and action selection}
Learned value functions have been used to improve expressive robot policies in several ways.
V-GPS~\cite{nakamoto2024steering} re-ranks sampled actions at test time using a learned value function.
More recently, RECAP~\cite{pistar06} uses learned value estimates to derive advantage signals for improving a VLA policy from heterogeneous robot experience.
SfBC~\cite{Chen2022OfflineRL} combines a multimodal generative behavior model with in-sample planning for offline reinforcement learning, enabling implicit stitching and value propagation across trajectories.
Our work builds on this in-sample planning principle for long-horizon coordination from mixed sub-task demonstrations, without explicit skill or sequencing variables, while improving the reliability of cross-demonstration value propagation.


\begin{figure*}[t]
\centerline{\includegraphics[width=0.8\textwidth]{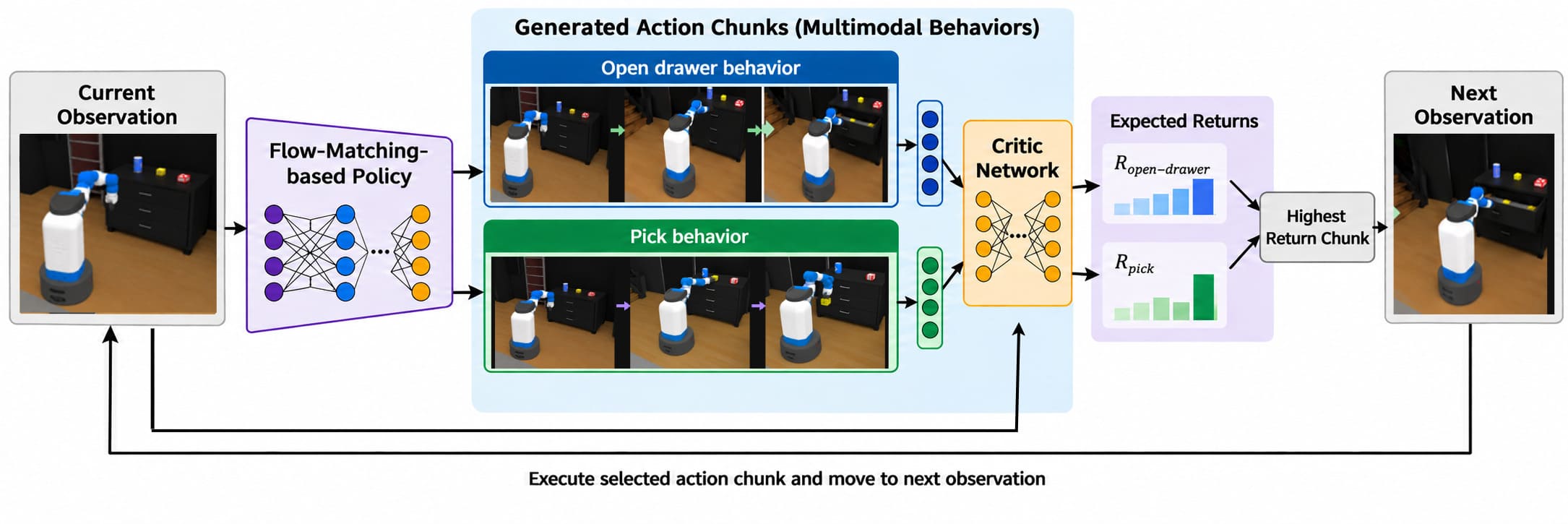}}
\caption{
The Flow Matching policy samples multiple candidate action chunks from a learned multimodal distribution conditioned on the current observation context.
At overlapping contexts, these candidates may correspond to different behavior continuations.
The critic scores each candidate according to the task objective, and the action chunk with the highest expected return is selected for execution.
}
\label{fig:framework}
\vspace{-0.5cm}
\end{figure*}

\section{Problem Formulation}
\label{sec:problem_formulation}

We consider long-horizon robotic rearrangement in a partially observed setting. 
The initial object position and desired target position are provided, but the object may not be directly accessible, e.g., enclosed inside a receptacle, and no environment map or obstacle locations are given. 
The task objective is to place the object at the target location. 
At each time step, the robot gets as observation the relative initial object position, the relative target position, arm joint positions, a binary gripper-holding indicator, and depth images from sensors mounted on the robot's head and end effector. 
The action consists of base linear and angular velocities, incremental arm joint commands, and a binary gripper command.

Unlike prior work that assumes complete task trajectories or explicit skills and coordination rules, we consider a more restrictive learning setting where the agent is given only demonstrations of individual sub-tasks, such as navigation, receptacle opening, picking, and placing. 
Each demonstration is a sequence of observations and actions without behavior identity labels or full-task ordering annotations.
The goal is to learn a policy that maps current observations to robot actions and adaptively coordinates the behaviors present in the demonstration data according to the task objective. 
Because the training data contains only sub-task demonstrations rather than complete task executions, the main challenge is to learn not only the individual behaviors but also how to coordinate them toward the sparse rearrangement objective.


\section{Our Approach}

We propose a framework that combines a multimodal generative policy with a learned critic for adaptive coordination, as illustrated in Fig.~\ref{fig:framework}. Demonstrations from different sub-tasks contain overlapping observations with different corresponding actions, as shown in Fig.~\ref{fig:cover_fig}. To model this multimodality, we use a conditional Flow Matching policy \cite{lipman2023flow}, which generates multiple candidate actions corresponding to different plausible behaviors. The critic evaluates these candidates under the task objective and selects the one with the highest value. Repeating this process allows the robot to coordinate different behaviors without complete task trajectories or hand-crafted coordination rules.

\subsection{Multi-Modal Flow Matching Policy}

We train a conditional Flow Matching policy with a transformer backbone on the mixed sub-task demonstrations. Each demonstration trajectory is converted into context--chunk samples $(s_n,a_n)$, where $s_n$ denotes a context of the previous $C$ observations and $a_n$ the corresponding chunk of $H$ future actions. Using action chunks improves temporal consistency and avoids querying the generative policy at every control step.

For each training sample, let $a_n$ denote the expert action chunk and let $x_0 \sim \mathcal{N}(0,I)$ be Gaussian noise of the same shape. We construct the interpolated input
\begin{equation} \small
x_\tau = (1-\tau)x_0 + \tau a_n, \qquad \tau \sim \mathcal{U}(0,1),
\label{flow_matching_noising_equation}
\end{equation}
with target velocity
\begin{equation} \small
v^\star = a_n - x_0.
\label{flow_matching_target_velocity_vector_calculation}
\end{equation}
The observation context $s_n$ is encoded into condition tokens using visual encoders for the depth images and a linear projection for the non-visual observations. Conditioned on these tokens, the policy predicts the velocity field from $x_\tau$ and $\tau$, and is trained by minimizing
\begin{equation} \small
\mathcal{L}_{\mathrm{FM}}
=
\mathbb{E}_{(s_n,a_n)\sim \mathcal{D},\,x_0,\tau}
\left[
\left\|
f_\theta(x_\tau,\tau,s_n) - v^\star
\right\|_2^2
\right].
\label{flow_matching_loss_function}
\end{equation}

At inference, we initialize $K$ action chunks from Gaussian noise and denoise them with the context-conditioned policy. The generated chunks are then evaluated by the critic, and the chunk with the highest predicted value is selected. We execute the first $E$ actions in the selected chunk and then query the policy again.

\subsection{Critic Network for Adaptive Behavior Selection and Coordination}

The critic $Q(s_n,a_n)$ estimates the expected return, that is, the discounted sum of future rewards, of executing an action chunk $a_n$ at observation context $s_n$ with respect to the task objective. Training the critic requires return targets for context--chunk pairs in the dataset. However, reward is sparse and only available at the final target state, so only demonstrations that directly reach this state receive non-zero return targets. To propagate reward information across different demonstrations, we adopt the in-sample planning method proposed in \cite{Chen2022OfflineRL} and modify it to our context--chunk setting.

The return target of each context--chunk sample is updated as
\begin{equation} \small
\label{eq:in-sample-planning}
\begin{aligned}
R_n^{(i)} &= r_n + \gamma \max \left( R_{n^{+}}^{(i)},\,
V_{n^{+}}^{(i-1)} \right),\\
V_{n^{+}}^{(i-1)} &:= \mathbb{E}_{a \sim \pi(\cdot \mid s_{n^{+}})}
Q_{\phi}(s_{n^{+}}, a).
\end{aligned}
\end{equation}
The critic is trained by minimizing the squared error between the predicted value $Q_{\phi}(s_n,a_n)$ and the updated return target~$R_n$. Here, $R_n$ is the updated return target for the current context--chunk sample, $r_n$ is the reward of action chunk $a_n$, obtained by summing the rewards over its $H$ actions, $R_{n^{+}}$ is the return of the next applicable context--chunk sample in the same trajectory, $V_{n^{+}}$ is the value of the next observation context $s_{n^{+}}$, which is aligned with the end of the current action chunk $a_n$, and $i$ denotes the in-sample planning iteration.

This update propagates value in two ways: $R_{n^{+}}$ propagates return backward within the same demonstration, while $V_{n^{+}}$ enables value transfer across demonstrations. At overlapping contexts, the multimodal policy can generate high-value action chunks consistent with another demonstration, increasing $V_{n^{+}}$ and consequently the return target of the current dataset chunk $(s_n,a_n)$.

\begin{figure*}[t]
\centering
\subfloat[Placing-like behavior.\label{fig:placing_skill_demonstartion_returns}]{%
\begin{minipage}[t]{0.22\textwidth}
\centering
\includegraphics[width=\textwidth,height=5.2cm,keepaspectratio]{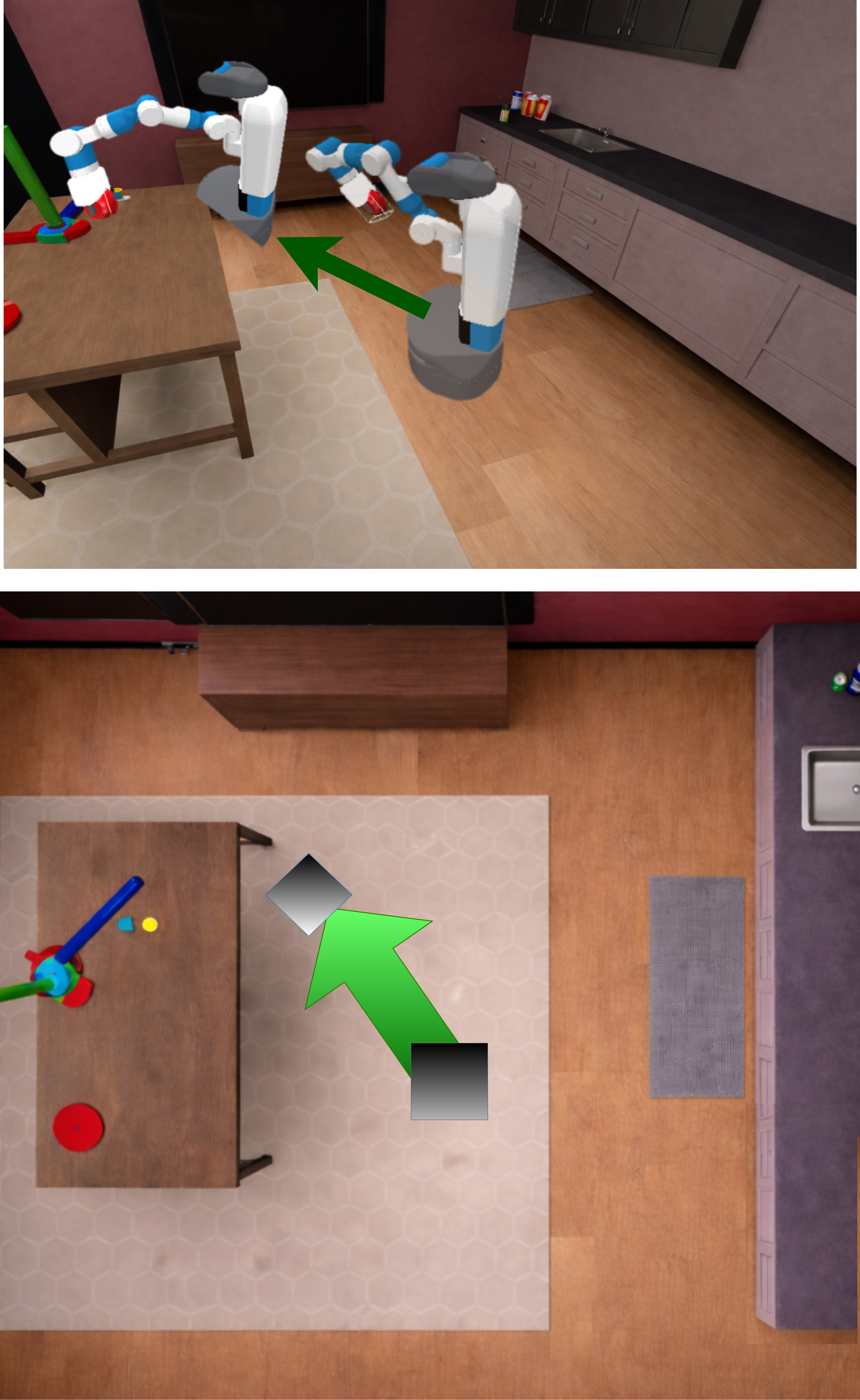}
\end{minipage}}\hfill
\subfloat[Navigation-like behavior.\label{fig:navigation_skill_demonstration_returns}]{%
\begin{minipage}[t]{0.22\textwidth}
\centering
\includegraphics[width=\textwidth,height=5.2cm,keepaspectratio]{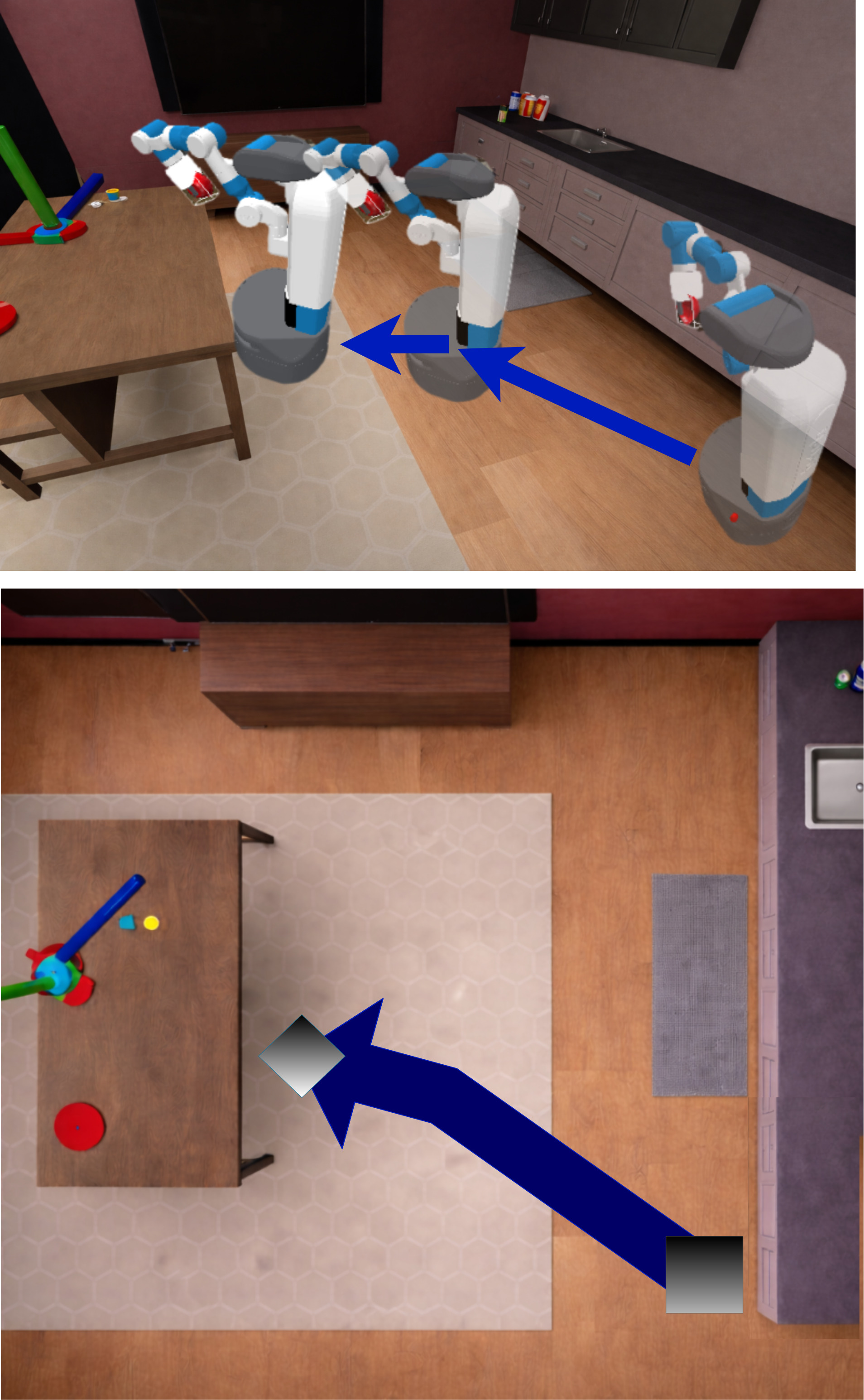}
\end{minipage}}\hfill
\subfloat[Returns without in-sample planning.\label{fig:returns_propagation_without_insample_planning}]{%
\begin{minipage}[t]{0.22\textwidth}
\centering
\includegraphics[width=\textwidth,height=5.2cm,keepaspectratio]{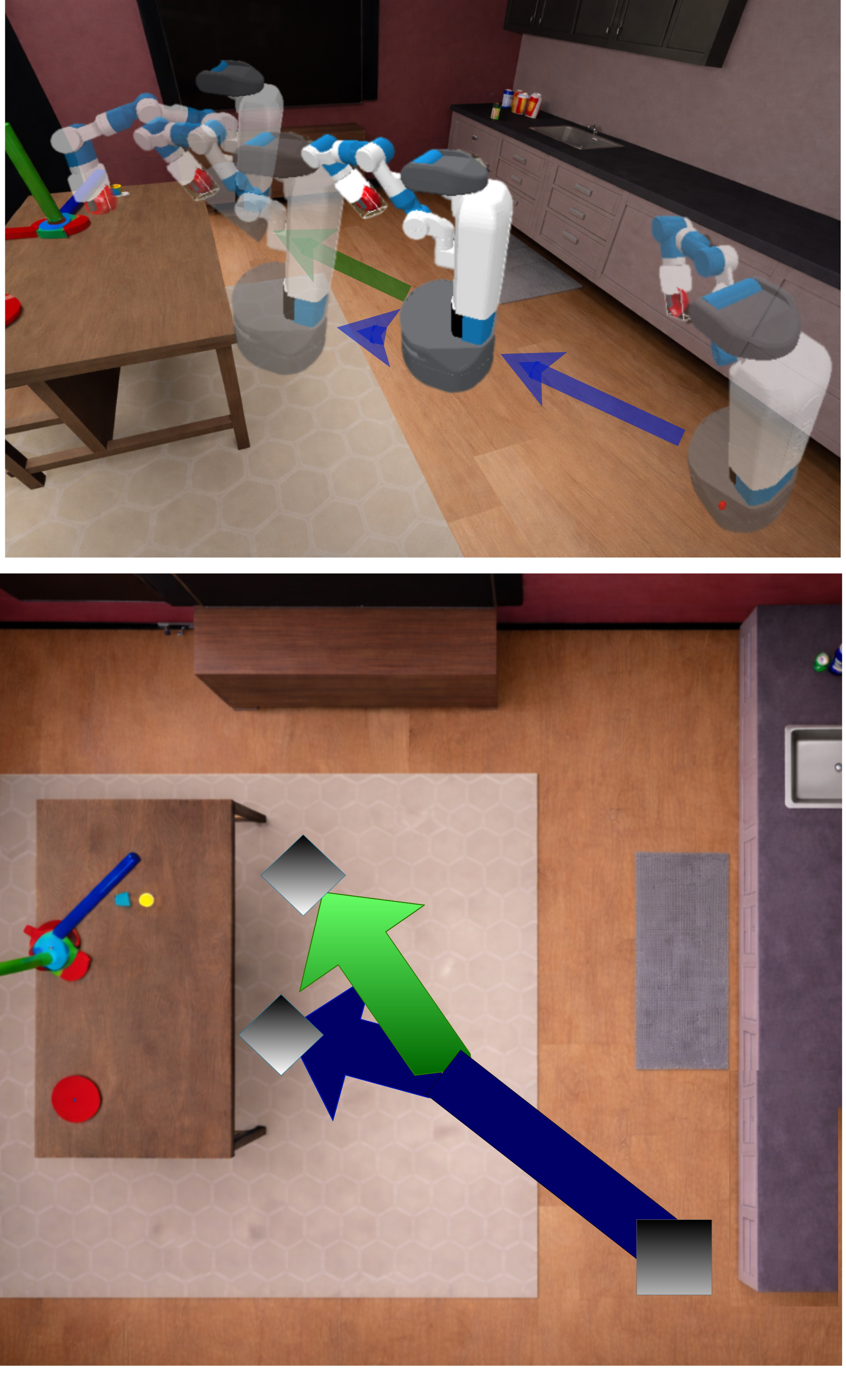}
\end{minipage}}\hfill
\subfloat[Returns with in-sample planning.\label{fig:returns_propagation_with_insample_planning}]{%
\begin{minipage}[t]{0.22\textwidth}
\centering
\includegraphics[width=\textwidth,height=5.2cm,keepaspectratio]{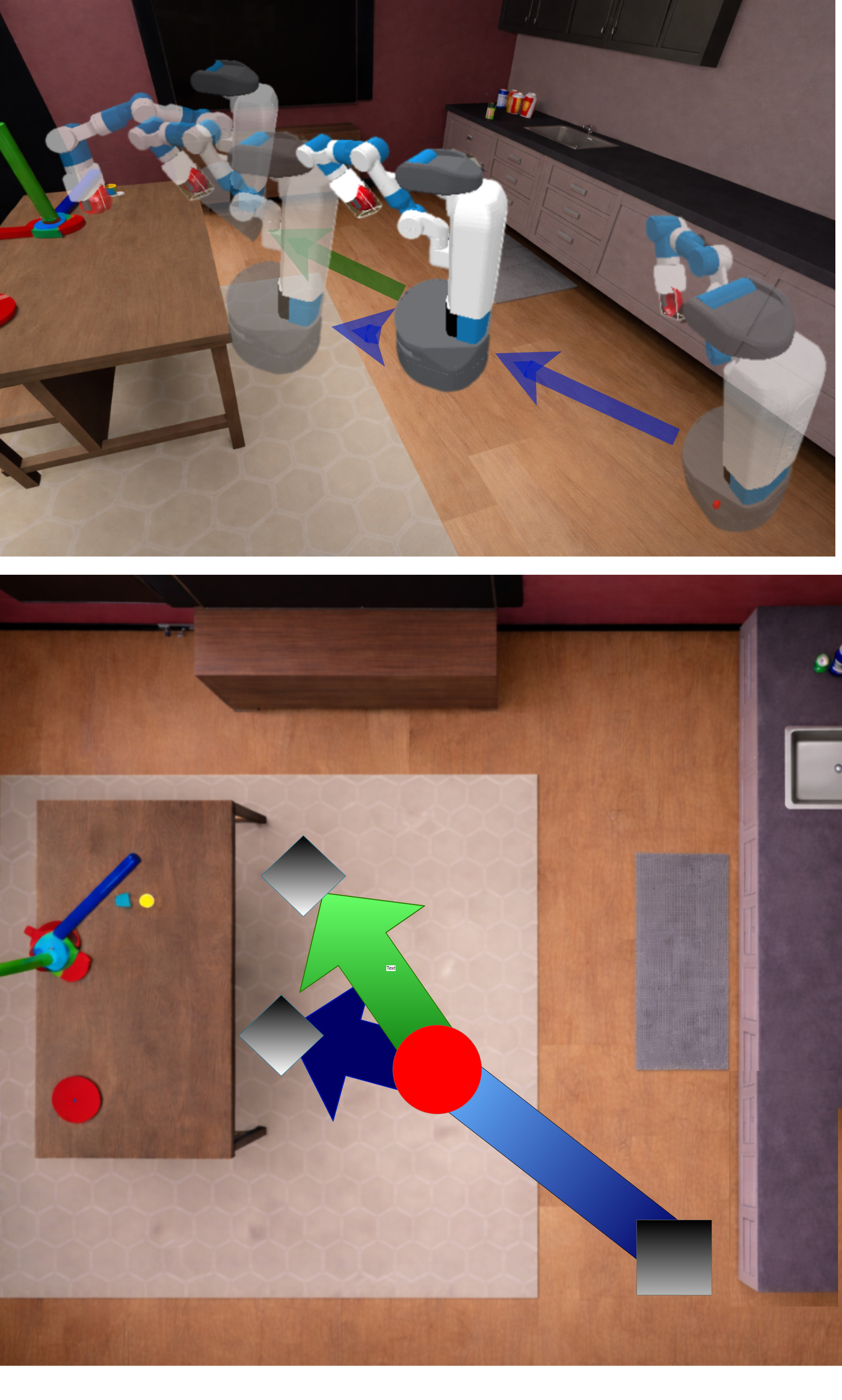}
\end{minipage}}
\caption{a) The placing behavior demonstration directly accesses the sparse reward and receives non-zero returns. b) The navigation-like demonstration with no access to the sparse reward has zero return values. c) Without in-sample planning, these zero returns remain unchanged along the navigation trajectory. d) With in-sample planning, overlapping contexts allow the Flow Matching policy to generate high-value placing actions within the navigation context, inducing non-zero returns that are then propagated backward through the navigation trajectory. Repeating this process over training iterations gradually spreads value information across overlapping demonstrations.
}
\label{fig:effect_of_using_insample_planning_for_critic_training}
\vspace{-0.5cm}
\end{figure*}

A concrete rearrangement example is illustrated in Fig.~\ref{fig:effect_of_using_insample_planning_for_critic_training}. The placing-like demonstration directly accesses the sparse reward and therefore has non-zero returns, whereas the navigation-like demonstration has zero return values under standard return propagation. Brighter colors represent a relatively higher return value. At overlapping observation contexts, the Flow Matching policy generates placing-like and navigation-like actions, implicitly stitching the demonstrations together. With in-sample planning, the high-return placing-like actions induce a high next-context value for the navigation-like demonstration, which is then propagated backward to earlier steps and produces non-zero returns. Repeating this procedure over in-sample planning iterations gradually propagates sparse reward information through reachable chains of overlapping demonstration trajectories.

However, this cross-demonstration bootstrap can be unreliable when the Flow Matching policy generates an action chunk that is weakly supported at the next context $s_{n^{+}}$. An overestimated critic value for such a continuation can increase $V_{n^{+}}$ and consequently inflate the return target of the current dataset chunk through Eq.~\ref{eq:in-sample-planning}. We therefore modify the computation of the next-context value $V_{n^{+}}$. Instead of the softmax-weighted aggregation used in~\cite{Chen2022OfflineRL}, we use an uncertainty-weighted average over the $K$ candidate action chunks generated by the Flow Matching policy at the next observation context $s_{n^{+}}$. For each candidate $a_k$, the critic is evaluated multiple times using Monte Carlo dropout to obtain the variance $\sigma_k^2$. The weight of each candidate and the next-context value are computed as
\begin{equation}  \small
w_k =
\frac{(\sigma_k^2)^{-1}}
{\sum_{j=1}^{K}(\sigma_j^2)^{-1}},
\qquad
V_{n^{+}}^{(i-1)}
=
\sum_{k=1}^{K} w_k \, Q_\phi(s_{n^{+}}, a_k).
\end{equation}
The uncertainty weighting acts on the next-context value used for bootstrapping. Generated action chunks at the next context with high critic variance receive smaller weights in $V_{n^+}$, reducing the influence of uncertain value estimates on the return target $R_n$ of the current dataset chunk $(s_n,a_n)$. Thus, it does not directly penalize particular behavior modes, but makes cross-demonstration value propagation less sensitive to unreliable critic estimates.

\section{Experiments}

Our experiments are organized around three questions corresponding to the three contributions outlined in Sec.~\ref{sec:introduction}.
First, can implicit behavior coordination be learned directly from separate sub-task demonstrations, without behavior identity labels, complete-task demonstrations, or an explicit sequencing policy, and how does it compare with explicit-skill and full-task imitation baselines?
We study this question in Secs.~\ref{sec:main_comparison} and~\ref{sec:data_architecture}.
Second, does overlap-induced multimodality enable coordination across demonstrations, and does uncertainty-aware value propagation improve the reliability of this process?
We examine these mechanisms through component comparisons and controlled reductions in cross-behavior overlap in Secs.~\ref{sec:main_comparison} and~\ref{sec:overlap}.
Third, how robust and flexible is implicit behavior coordination under suboptimal demonstrations with reduced cross-behavior overlap, larger mixed behavior repertoires, longer chained objectives, and real-world failures and recovery?
We evaluate these settings in Secs.~\ref{sec:overlap}--\ref{sec:realworld}.

\begin{table*}[t]
\centering
\resizebox{0.7\textwidth}{!}{
\begin{tabular}{l c c c}
\toprule
\textbf{Paradigm / Method}
& \textbf{Nav-Place}
& \textbf{Nav-Pick-Nav-Place}
& \textbf{Nav-Open-PickDr-Nav-Place} \\
\midrule

\multicolumn{4}{l}{\textit{Explicit-skill coordination}} \\

Oracle Planner + RL Skills
& $100.0\pm0.0$
& $95.7\pm0.5$
& $93.0\pm2.2$ \\

Oracle Planner + BC Skills
& $97.0\pm0.8$
& $80.3\pm1.2$
& $70.0\pm4.5$ \\

VLM Planner + RL Skills
& $96.5\pm0.7$
& $86.5\pm3.5$
& $35.0\pm1.4$ \\

VLM Planner + RL Skills + Privileged State
& $96.5\pm0.7$
& $96.5\pm0.7$
& $88.5\pm2.1$ \\

\midrule

\multicolumn{4}{l}{\textit{Task-specific full-task imitation}} \\

ST$_{\text{Nav-Place}}$
& $91.0\pm1.0$
& --
& -- \\

ST$_{\text{Nav-Pick-Nav-Place}}$
& $69.0\pm2.6$
& $65.6\pm1.2$
& -- \\

ST$_{\text{Nav-Open-PickDr-Nav-Place}}$
& $76.5\pm0.7$
& $2.5\pm2.1$
& $58.5\pm2.0$ \\

\midrule

\multicolumn{4}{l}{
\textit{Implicit coordination from sub-task demonstrations}
} \\

FM
& $88.5\pm2.1$
& $72.0\pm4.2$
& $64.0\pm8.6$ \\

FM + Critic
& $90.0\pm1.4$
& $75.0\pm1.4$
& $55.5\pm2.5$ \\

FM + Critic + ISP
& $90.5\pm0.7$
& $74.5\pm0.7$
& $61.3\pm1.7$ \\

\textbf{Ours}
& $89.3\pm2.0$
& $76.3\pm2.5$
& $68.5\pm2.4$ \\

\bottomrule
\end{tabular}
}
\caption{
Success rate (\%, mean $\pm$ std) on rearrangement tasks with increasing coordination complexity.
A dash (--) denotes configurations that were not applicable. ST and FM denote Skill Transformer and Flow Matching, respectively.
Our implicit-coordination approach achieves competitive performance without explicit skill sequencing or complete-task demonstrations and performs best among implicit-coordination variants on the most complex task.
}
\label{tab:main_results}
\vspace{-0.5cm}
\end{table*}

\subsection{Simulation Setup}
\label{sec:simulation_setup}

\subsubsection{Tasks} 
We use Habitat 2.0 with the ReplicaCAD dataset and the Fetch robot in home-like environments~\cite{hab_2_0} and evaluate three rearrangement tasks with increasing coordination complexity:
\textbf{Nav-Place}, where the robot already holds the object and must navigate to the target and place it;
\textbf{Nav-Pick-Nav-Place}, where the robot must navigate to the object, pick it, navigate to the target, and place it;
and \textbf{Nav-Open-PickDr-Nav-Place}, where the robot must navigate to a drawer, open it, pick the object from the drawer, navigate to the target, and place it.
We distinguish \textbf{Pick} from \textbf{PickDr}: after opening the drawer, the object's position may differ from its given initial position, requiring the agent to rely on depth observations to localize and pick it.
Task success is defined as placing the object at the target location within the maximum allowed number of steps.

\subsubsection{Demonstration Data}
We generate all simulation demonstrations in the training scenes used by M3~\cite{M3}, using pretrained M3 RL skill policies.
We consider two forms of demonstration data.
\textbf{Complete-task demonstrations} consist of 10,000 \textit{successful} skill-labeled trajectories per task, generated by the Oracle Planner + RL Skills system.
\textbf{Sub-task demonstrations} are collected separately for individual behavior modes, with 6,400 \textit{successful} demonstrations per mode, matching the number of episodes in the M3 training split.
The full sub-task pool contains six behavior modes: navigation to the pick position, navigation to the place position, picking, drawer opening, picking from a drawer, and placing.

\subsubsection{Compared Methods}
We compare three classes of approaches corresponding to different paradigms.

(i) Explicit-skill coordination:
\textbf{Oracle Planner + RL Skills} uses privileged environment information to select the correct skill sequence.
\textbf{Oracle Planner + BC Skills} uses the same oracle planner but executes behavior-cloned skill policies.
For learned explicit coordination, \textbf{VLM Planner + RL Skills} uses GPT-4o to select the next skill from a predefined skill library based on RGB-D observations, robot/task state, and skill initiation conditions.
\textbf{VLM Planner + RL Skills + Privileged State} additionally receives explicit task-relevant semantic state, such as whether the target object is located inside a drawer.

(ii) Task-specific full-task imitation:
\textbf{Skill Transformer~(ST)}~\cite{skill_transformer} jointly predicts a high-level skill and low-level actions from complete-task demonstrations.

(iii) Implicit coordination from sub-task demonstrations:
\textbf{FM} uses a shared Flow Matching policy without value guidance.
\textbf{FM + Critic} adds critic-guided candidate selection at inference time. The critic is trained using within-trajectory temporal credit assignment, where each target return is updated using only the return of the next context--chunk in the same trajectory.
\textbf{FM + Critic + ISP} further adapts in-sample planning (ISP)~\cite{Chen2022OfflineRL} to our context--chunk setting and uses Eq.~\ref{eq:in-sample-planning} to enable cross-demonstration value propagation, with softmax-weighted aggregation to compute the next-context value.
\textbf{Ours} replaces this softmax aggregation with uncertainty-aware weighting based on critic variance.

\subsubsection{Training and Supervision}
The RL-skill variants use the same pretrained M3 RL skill checkpoints and require no additional low-level policy training.
The BC-skill baseline trains separate behavior-cloned skill policies from the corresponding sub-task demonstrations.
ST follows its original training procedure and is trained separately for each task using the complete-task demonstrations described above.
All implicit-coordination variants train a single shared Flow Matching policy by mixing the included sub-task demonstrations without behavior identity labels.
Unless otherwise specified, our implicit-coordination variants use the full six-mode pool, corresponding to $6\times6{,}400$ demonstrations.
In task-specific studies, only the behavior modes required by the corresponding task are included.
For critic-based variants, the critic is trained using a sparse task reward indicating whether the object is placed at the target.
Study-specific modifications to the training data or supervision are described in the corresponding subsections.

\subsubsection{Evaluation Protocol}
For each task, we use a fixed set of 100 episodes from the M3~\cite{M3} validation split.
These episodes use the same scene layouts as the training split but different object locations.
Unless otherwise specified, all simulation methods are evaluated on the same 100 episodes per task with three random seeds.
Each episode has a 5,000-step limit, following~\cite{skill_transformer}.
Study-specific protocol changes are described in the corresponding subsections.

\subsubsection{Implementation Details}
At each decision step, our method samples and evaluates $K=20$ candidate action chunks.
This takes approximately 1.7\,s per decision. The values of $C$, $E$, and $H$ are 5, 10, and 20, respectively.

\subsection{Main Comparison across Different Paradigms}
\label{sec:main_comparison}

\textbf{Comparison to task-specific full-task imitation.}
Table~\ref{tab:main_results} first compares our method with task-specific full-task imitation.
On the task-matched evaluations, our method performs comparably to ST on Nav-Place and outperforms it on the two longer-horizon tasks, despite using only separate sub-task demonstrations without behavior identity labels or task-ordering supervision.
Many ST failures occur during picking and placing, where deterministic action prediction struggles with multimodal action distributions within behaviors.
We additionally evaluate ST policies on simpler task variants to probe cross-task transfer.
The particularly poor transfer of ST$_{\text{Nav-Open-PickDr-Nav-Place}}$ to Nav-Pick-Nav-Place is largely due to the standard Pick behavior being absent from its training task, which instead contains PickDr. 

\textbf{Comparison to explicit-skill coordination.}
The explicit-skill baselines provide a complementary comparison.
These methods assume a predefined skill library and an explicit skill-selection interface, while the oracle and privileged-state variants additionally use privileged task information.
Replacing the pretrained RL skills with behavior-cloned skills causes a clear performance drop under the same oracle planner, consistent with prior observations in Skill Transformer~\cite{skill_transformer}.
On the most complex Nav-Open-PickDr-Nav-Place task, our method nearly matches Oracle Planner + BC Skills despite not using an oracle skill sequence or explicit behavior identities.
On the longer-horizon tasks, the VLM planner performs substantially worse without privileged state, but improves markedly when task-relevant semantic state is provided, highlighting the importance of reliable state grounding for explicit skill selection.

\textbf{Ablation of implicit coordination.}
The implicit-coordination variants isolate the contribution of value guidance and uncertainty-aware ISP.
On Nav-Open-PickDr-Nav-Place, adding critic guidance alone does not improve over FM, while standard ISP partially recovers performance and our uncertainty-aware aggregation achieves the best result.
This suggests that value guidance alone is not sufficient, and that reliable value propagation across overlapping behaviors becomes important as coordination complexity increases.
The effect is illustrated in Fig.~\ref{fig:cover_fig}.
Around similar observation contexts near the drawer, the shared FM policy can generate action chunks corresponding to either drawer opening or picking.
Because picking is closer to the sparse task reward, standard ISP may assign high propagated value to picking-like continuations even when they are weakly supported at that context.
In contrast, drawer-opening continuations are better supported by the demonstrations and yield more consistent critic estimates.
Our uncertainty-aware aggregation downweights high-variance value estimates, reducing unreliable propagation and favoring better-supported behavior continuations.

Overall, these results position implicit coordination from separate sub-task demonstrations as a viable alternative to explicit skill sequencing and task-specific full-task imitation, with uncertainty-aware value propagation improving its reliability in complex overlapping contexts.

\begin{table}[t]
\centering
\scriptsize
\setlength{\tabcolsep}{2.5pt}
\begin{tabular}{l l c}
\toprule
\textbf{Method} & \textbf{Demonstrations} &
\textbf{Success (\%)}  \\
\midrule
ST & Complete-task         & $65.6\pm1.2$ \\
ST & Sub-task        & $1.5\pm0.7$ \\
\midrule
FM & Complete-task        & $69.5\pm0.7$ \\
FM & Sub-task        & $76.3\pm2.0$ \\
\textbf{Ours} & Sub-task & $78.5\pm2.1$ \\
\bottomrule
\end{tabular}
\caption{
Effect of policy architecture and demonstration collection on Nav--Pick--Nav--Place.
FM benefits from separate sub-task demonstrations, whereas ST relies strongly on complete-task trajectories; our value-guided variant achieves the highest success.
}
\label{tab:controlled_ablation}
\vspace{-0.5cm}
\end{table}

\subsection{Effect of Policy Architecture and Demonstration Collection}
\label{sec:data_architecture}

\textbf{Controlled setting.}
Table~\ref{tab:main_results} compares each paradigm under its native training interface and supervision.
Here, we instead control the task and training data to isolate the effects of policy architecture and demonstration organization.
We focus on Nav-Pick-Nav-Place and use task-specific training data.
The complete-task variants of ST and FM are trained on the same 10,000 successful complete-task demonstrations.
For separate sub-task training, we include only the four behavior modes required by this task, corresponding to $4\times6{,}400$ demonstrations.

\textbf{Effect of policy architecture.}
Table~\ref{tab:controlled_ablation} studies policy architecture and demonstration collection on Nav--Pick--Nav--Place.
ST and FM with complete trajectories are trained on the same data.
FM performs better, suggesting that modeling multimodal action distributions is beneficial even within individual behaviors.
In contrast, ST performance collapses with separate sub-task demonstrations, showing its reliance on complete trajectories to infer behavior transitions.

\textbf{Effect of demonstration collection.}
FM and our method use a task-specific mixed pool containing only the sub-tasks required for Nav--Pick--Nav--Place.
These separate demonstrations provide broader state coverage than complete trajectories, which follow relatively fixed pick-to-place paths, improving robustness to covariate shift and compounding errors.
FM and our method perform similarly because these demonstrations are successful and largely transition-compatible, leaving limited need for task-directed behavior selection.

Overall, these results support separate sub-task demonstrations as a practical training interface.

\begin{table}[t]
\centering
\footnotesize
\setlength{\tabcolsep}{3.2pt}
\resizebox{\linewidth}{!}{
\begin{tabular}{@{}l c c c@{}}
\toprule
\textbf{Metric} &
\textbf{0\% Incorrect Nav.} &
\textbf{50\% Incorrect Nav.} &
\textbf{74\% Incorrect Nav.} \\
\midrule
Context overlap (\%) &
5.82 &
4.52 &
3.10 \\

Trajectory overlap (\%) &
25.67 &
19.70 &
15.14 \\
\midrule

FM success rate (\%) &
$83.0\pm4.2$ &
$52.0\pm2.8$ &
$45.0\pm1.4$\\

\textbf{Ours} success rate (\%) &
$81.0\pm4.2$ &
$57.0\pm1.4$ &
$53.0\pm1.4$ \\
\bottomrule
\end{tabular}
}
\caption{
Effect of reduced cross-behavior overlap on Nav-Pick-Nav-Place.
Increasing the fraction of incorrect navigation demonstrations reduces both context and trajectory overlap and degrades FM substantially, while our method remains more robust.
}
\label{tab:overlap_sensitivity}
\vspace{-0.5cm}
\end{table}

\subsection{Effect of Suboptimal Demos and Behavior Overlap}
\label{sec:overlap}

\textbf{Controlled setting.}
Because cross-behavior overlap is induced by the states visited by the demonstrations, directly varying overlap while holding the demonstration state distribution fixed is not generally possible.
We therefore start from the task-specific Nav-Pick-Nav-Place sub-task pool used in Sec.~\ref{sec:data_architecture} and construct three training sets containing $0\%$, $50\%$, and $74\%$ incorrect navigation trajectories.
Incorrect navigation trajectories terminate at positions unsuitable for initiating the subsequent picking or placing behavior, thereby reducing transition compatibility between successive behaviors.
All other sub-task demonstrations are kept unchanged.
Unlike the standard evaluation protocol, this study uses a fixed 50-episode validation subset whose scene layouts are represented across all three training sets, isolating the effect of cross-behavior overlap from scene coverage.
The absolute success rates in this table are thus not directly comparable to those in Table~\ref{tab:controlled_ablation}.

\textbf{Overlap measurement.}
To quantify overlap between successive behaviors, we train a VAE on the demonstration data and compute cosine similarity between latent observation-context representations.
At a threshold of $0.8$, context overlap measures the fraction of contexts matching a context from the successive behavior, while trajectory overlap measures the fraction of trajectories containing at least one such matching context.
Since trajectory lengths and the number of valid contexts differ, we report normalized overlap percentages.

\textbf{Effect of reduced overlap.}
As shown in Table~\ref{tab:overlap_sensitivity}, increasing the proportion of incorrect navigation trajectories progressively reduces both context- and trajectory-level overlap.
As overlap decreases, FM success drops, whereas our method degrades less.
Notably, the measured context overlap is relatively sparse even in the clean setting, yet it is sufficient to support cross-behavior coordination.
Moreover, performance degrades gradually rather than collapsing as overlap is reduced, suggesting that implicit coordination does not require dense coverage of behavior transitions.
This indicates that FM alone becomes increasingly sensitive to reduced transition compatibility, while critic-guided selection improves robustness by favoring task-compatible action continuations.

\begin{table}[t]
\centering
\resizebox{\linewidth}{!}{
\begin{tabular}{l c c c}
\toprule
\textbf{Task} & \textbf{Behaviors} & \textbf{Task-Specific Agent} & \textbf{Full-Repertoire Agent} \\
\midrule
Nav-Place & 2 & 87.5$\pm$2.1 & 89.3$\pm$2.0 \\
Nav-Pick-Nav-Place & 4 & 78.5$\pm$2.1 & 76.3$\pm$2.5 \\
Nav-Open-PickDr-Nav-Place & 5 & 69.5$\pm$0.7 & 68.5$\pm$2.4 \\
\bottomrule
\end{tabular}
}
\caption{
Effect of behavior repertoire size.
The full-repertoire agent maintains comparable performance across all tasks despite the larger mixed behavior pool.
}
\label{tab:behavior_scaling}
\vspace{-0.5cm}
\end{table}

\subsection{Effect of Behavior Repertoire Size}
\label{sec:repertoire}

\textbf{Controlled setting.}
We compare separately trained task-specific agents with a single full-repertoire agent (Ours).
Each task-specific agent is trained only on the behavior modes required by its corresponding task, following the task-specific setting in Sec.~\ref{sec:data_architecture}.
In contrast, the full-repertoire agent is trained once on all six available behavior modes and is evaluated on all three tasks without task-specific retraining.

\textbf{Effect of a larger behavior repertoire.}
Table~\ref{tab:behavior_scaling} shows that the full-repertoire agent performs similarly to the corresponding task-specific agents across all three tasks.
This indicates that adding behavior modes beyond those required by a particular task does not substantially degrade coordination performance.
More importantly, a single agent can select task-relevant behavior continuations from a larger mixed demonstration pool without behavior identity labels or task-specific coordination logic.

\subsection{Effect of Task Horizon with Chained Targets}
\label{sec:chained_targets}

\textbf{Controlled setting.}
We reuse the agents from the standard Nav-Pick-Nav-Place evaluation without retraining.
Each chained-target episode starts from the same 100 validation episodes used in the standard evaluation.
After a successful placement, a new target is issued and execution continues without resetting the environment, for up to five sequential targets.
This increases the number of required behavior transitions and compounds execution errors over a longer horizon.
We use Nav-Pick-Nav-Place because it naturally supports repeated object-target rearrangement goals and follows the tidy-house-style sequential rearrangement setting used in M3.

\textbf{Effect of longer task horizons.}
Table~\ref{tab:chained_target_results} reports cumulative success through each chained target.
Performance decreases for all methods as the number of targets increases, reflecting longer horizons and accumulated execution errors.
The oracle-planner baselines remain strongest, while ST degrades much more sharply than our method as additional targets are chained.
Without retraining, our implicit-coordination agent maintains substantially better performance than task-specific full-task imitation over longer horizons, suggesting greater resilience to repeated behavior transitions and accumulated errors.

\begin{table}[t]
\centering
\resizebox{\linewidth}{!}{
\begin{tabular}{l c c c c c}
\toprule
\textbf{Paradigm / Method}
& \textbf{1 Target}
& \textbf{2 Targets} 
& \textbf{3 Targets} 
& \textbf{4 Targets} 
& \textbf{5 Targets}\\
\midrule
\multicolumn{6}{l}{\textit{Explicit-skill coordination}} \\
Oracle Planner + RL Skills 
& 95.6$\pm$0.5 & 88.0$\pm$1.7 & 78.0$\pm$3.6 & 68.6$\pm$2.8 & 62.0$\pm$1.7 \\
Oracle Planner + BC Skills 
& 80.5$\pm$0.7 & 69.5$\pm$0.7 & 61.0$\pm$1.4 & 49.0$\pm$4.2 & 42.0$\pm$5.0 \\
\midrule
\multicolumn{6}{l}{\textit{Task-specific full-task imitation}} \\
ST$_{\text{Nav-Pick-Nav-Place}}$
& 65.0$\pm$1.4 & 23.5$\pm$2.1 & 8.0$\pm$2.8 & 2.0$\pm$1.4 & 0.5$\pm$0.7 \\
\midrule
\multicolumn{6}{l}{\textit{Implicit coordination from sub-task demonstrations}} \\
\textbf{Ours}
& 76.0$\pm$2.8 & 62.5$\pm$4.9 & 50.5$\pm$0.7 & 43.5$\pm$3.5 & 32.0$\pm$4.2 \\
\bottomrule
\end{tabular}
}
\caption{
Effect of increasing task horizon on chained-target Nav-Pick-Nav-Place.
Each episode continues across multiple sequential object-target goals without resetting the environment.
We report the cumulative success rate (\%) through each target.
Without retraining, our implicit-coordination agent degrades more gracefully than task-specific full-task imitation as the task horizon increases.
}
\label{tab:chained_target_results}
\vspace{-0.5cm}
\end{table}

\subsection{Real-World Validation}
\label{sec:realworld}

\begin{figure}[t]
    \centering
    \subfloat[Initial position.]{
        \includegraphics[
            width=0.4\linewidth,
            angle=-90,
            origin=c
        ]{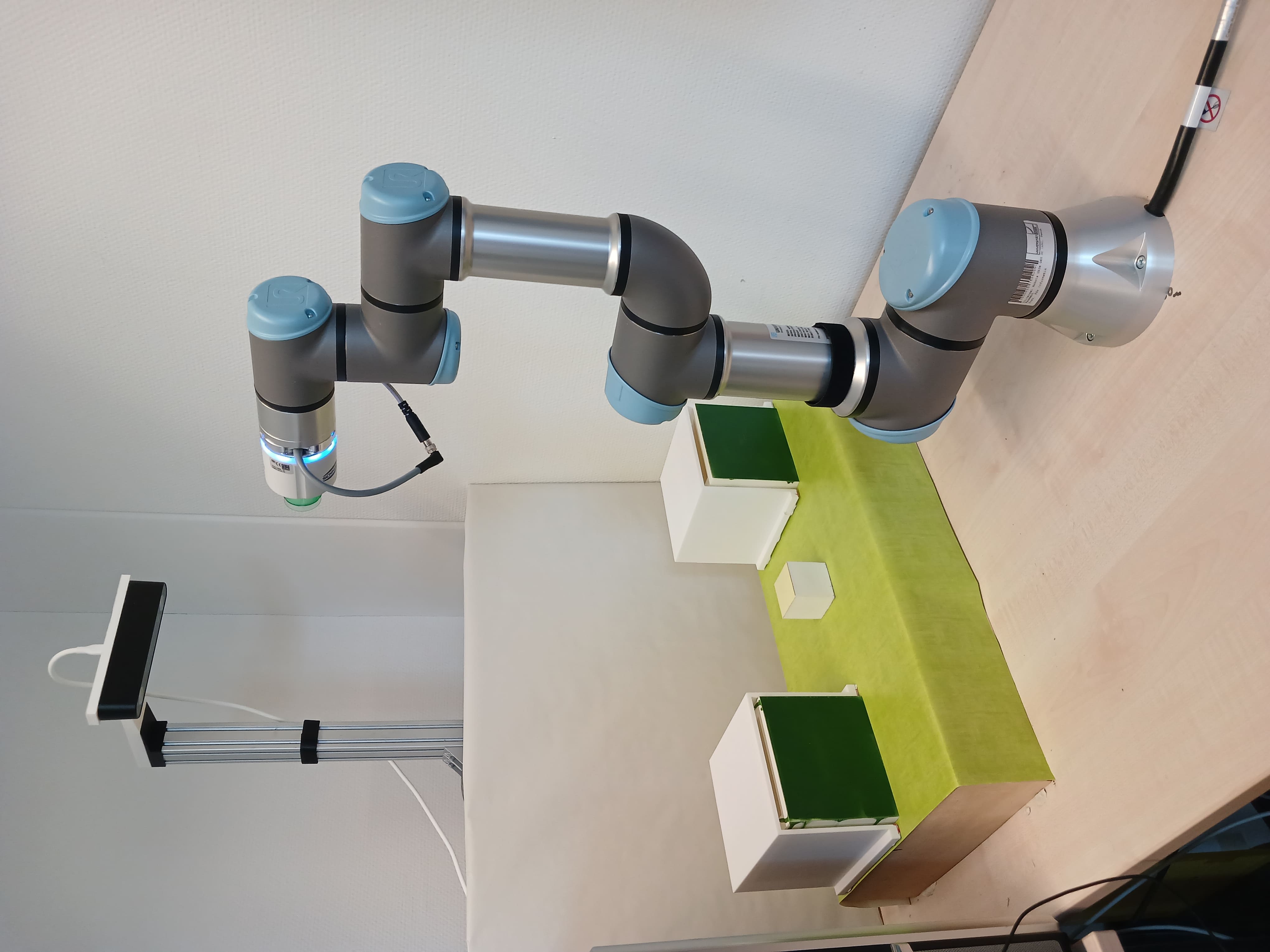}
        \label{fig:subfig_initial_pos_real_world}
    }
    \hfill
    \subfloat[Open left.]{
        \includegraphics[
            width=0.4\linewidth,
            angle=-90,
            origin=c
        ]{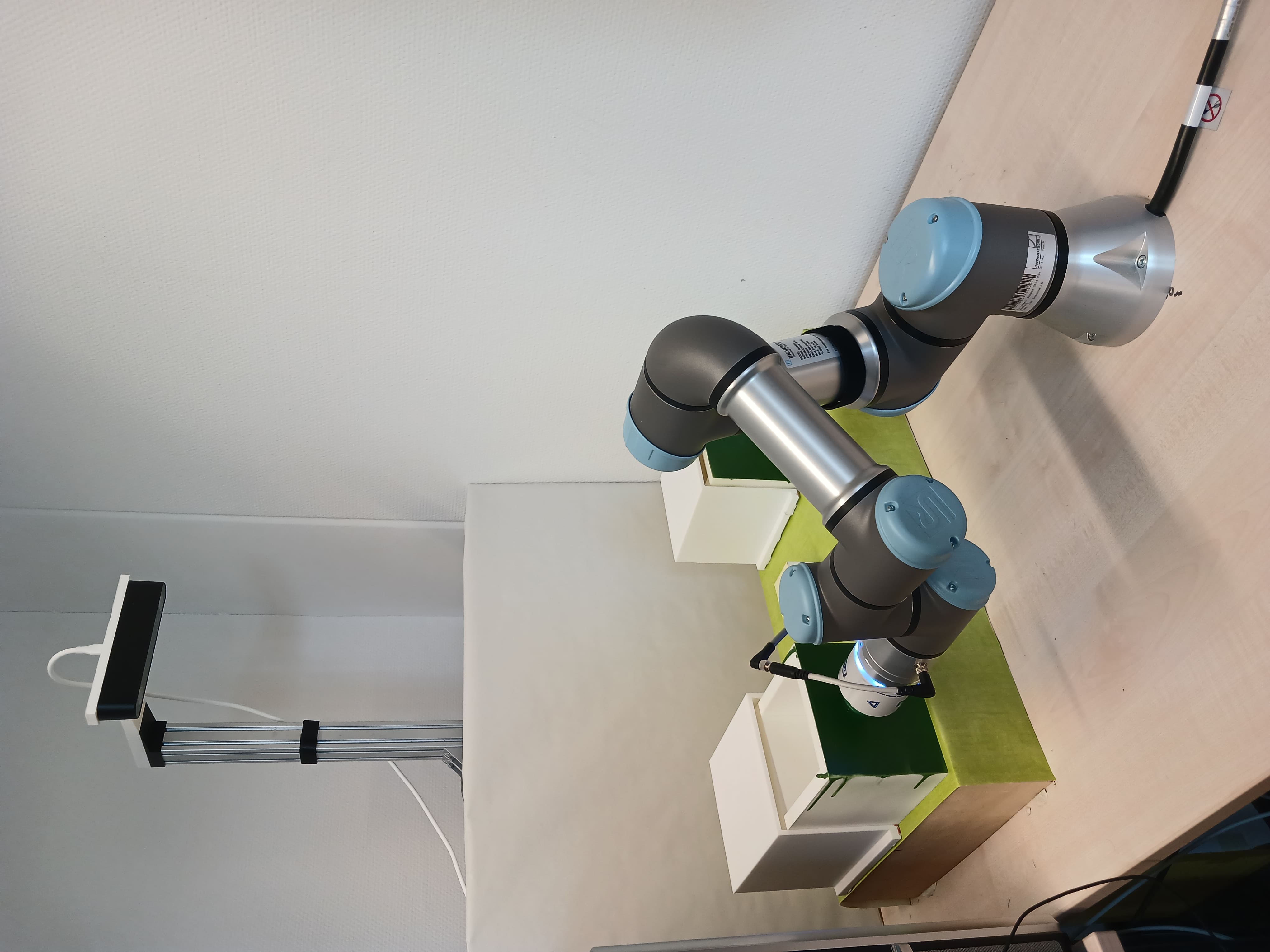}
        \label{fig:open_left_real_world}
    }
    \hfill
    \subfloat[Place left.]{
        \includegraphics[
            width=0.4\linewidth,
            angle=-90,
            origin=c
        ]{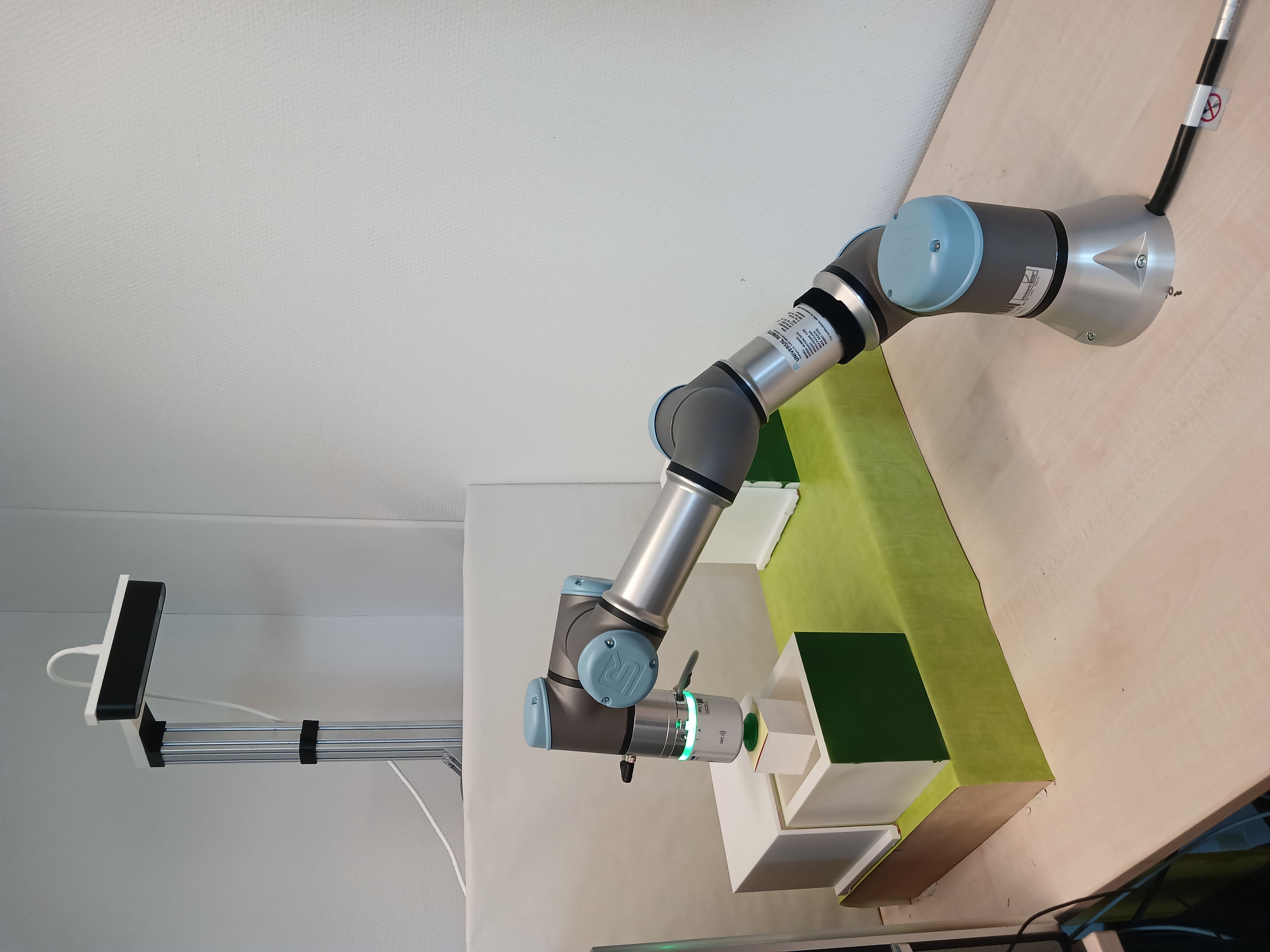}
        \label{fig:place_left_real_world}
    }
    \caption{
Real-world rearrangement setup.
The robot starts from configuration (a) and must first open the left drawer (b), then pick the object and place it in the left drawer (c).
}
    \label{fig:real_world_initial_and_final_pos}
    \vspace{-0.2cm}
\end{figure}

\textbf{Real-world setting.}
We evaluate our method on the tabletop rearrangement task shown in Fig.~\ref{fig:real_world_initial_and_final_pos}, where the objective is to place the target object in the left drawer.
We train a separate real-world agent for this task rather than transferring the simulation policy, so that the evaluation isolates behavior coordination from sim-to-real perception and dynamics gaps.
Unlike the simulation setup, the policy receives no explicit object target position or desired place location.
Its observations consist only of depth images from the top-down camera shown in Fig. \ref{fig:real_world_initial_and_final_pos}, robot joint positions, and the object-holding state.
Thus, the left-drawer objective is specified only through the task reward rather than being provided as an explicit goal input.

\textbf{Demonstrations and evaluation.}
We collect 50 real-world demonstrations spanning five behavior modes: \textit{opening the left drawer, opening the right drawer, picking, placing in the left drawer}, and \textit{placing in the right drawer}.
The demonstrations are mixed to train a single shared Flow Matching policy without behavior identity labels.
For critic-based coordination, a sparse reward is provided only when the object is successfully placed in the left drawer.
We evaluate FM and our full method over 20 real-world trials under the same task objective.

\textbf{Task-directed behavior selection.}
A key ambiguity arises because opening the left and right drawers share similar initial observation contexts.
Without explicit semantic or target-position information, the FM policy can generate action chunks corresponding to either behavior.
FM succeeds in 10/20 trials, whereas our method succeeds in 19/20, with the single failure occurring during picking.
This shows that task value can disambiguate plausible behavior continuations even when the task-relevant target is not explicitly provided to the policy.

\textbf{Value propagation and recovery.}
To examine how value propagates through suboptimal and recovery behaviors, we replaced 30 successful open left drawer trajectories with 10 recoverable and 20 non-recoverable failed trajectories. After failing to open the drawer, recoverable trajectories terminate at contexts from which task-relevant behavior can resume, whereas non-recoverable trajectories terminate at contexts that do not support continuation toward the task objective.
Table~\ref{tab:critic_values_real} shows how the sparse task reward propagates backward through the demonstration pool.
Value first appears for Place left, where the task reward is observed directly, and then propagates to Pick and Open left.
With additional ISP iterations, value also reaches recovery trajectories that reconnect to task-relevant contexts.
Recoverable trajectories receive higher values than failed, non-recoverable opening trajectories because they reconnect to states from which successful behavior continuations remain available.
Non-recoverable trajectories can nevertheless acquire moderate value because their early portions overlap with successful or recoverable trajectories before diverging.
In contrast, unrelated behaviors such as Open right and Place right remain near zero because they do not connect to the rewarded task objective.

\begin{table}[t]
\centering
\scriptsize
\setlength{\tabcolsep}{2.0pt}
\begin{tabular}{lccccc}
\toprule
\textbf{Behavior} &
\textbf{Iter. 1} &
\textbf{Iter. 2} &
\textbf{Iter. 3} &
\textbf{Iter. 5} &
\textbf{Iter. 10} \\
\midrule
Place left            & 91.8 & 91.7 & 92.3 & 92.8 & 92.7 \\
Pick        & 0.0 & 52.7 & 59.1 & 60.2 & 62.9 \\
Open left             & 0.0 & 0.0 & 7.2 & 44.5 & 48.5 \\
Recovery               & 0.0 & 0.0 & 0.0 & 0.2 & 33.3 \\
Failed open left      & 0.0 & 0.0 & 0.0 & 0.1 & 22.0 \\
Open right              & 0.0 & 0.0 & 0.0 & 0.1 & 0.2 \\
Place right             & 0.0 & 0.0 & 0.0 & 0.1 & 0.1 \\
\bottomrule
\end{tabular}
\caption{
Median critic values for real-world behaviors across in-sample planning iterations.
Sparse task value propagates backward from successful placement to preceding task-relevant and recoverable behaviors, while unrelated behaviors remain near zero.
}
\label{tab:critic_values_real}
\vspace{-0.5cm}
\end{table}

\section{Conclusion}
\label{sec:conclusion}

We presented implicit behavior coordination, a formulation for long-horizon rearrangement from separately collected sub-task demonstrations without behavior identity labels, complete-task demonstrations, or an explicit sequencing policy.
The approach preserves overlap-induced multimodality in a shared generative policy and uses task value to coordinate among plausible behavior continuations.
Across simulation and real-world experiments, it achieves competitive performance relative to explicit-skill coordination and task-specific full-task imitation, while remaining effective under reduced overlap, larger behavior mixtures, longer chained objectives, and execution failures.
These results suggest that useful long-horizon coordination can emerge from shared observation contexts and value-guided selection without explicitly representing behavior identities or sequencing interfaces.
Future work may explore more diverse environments and richer task conditioning for real-world rearrangement.


\bibliographystyle{IEEEtran}
\bibliography{references}

\end{document}